\documentclass[lettersize,journal]{IEEEtran}
\usepackage{amsmath,amsfonts}
\usepackage{algorithmic}
\usepackage{algorithm}
\usepackage{array}
\usepackage[caption=false,font=normalsize,labelfont=sf,textfont=sf]{subfig}
\usepackage{textcomp}
\usepackage{stfloats}
\usepackage{url}
\usepackage{verbatim}
\usepackage{graphicx}
\usepackage{cite}
\usepackage{xcolor}
\usepackage{cleveref}
\usepackage[inline]{enumitem}
\usepackage{booktabs}
\usepackage{eso-pic}
\usepackage{url}
\AddToShipoutPictureBG*{
  \AtPageUpperLeft{%
    \put(0,-40){\raisebox{15pt}{\makebox[\paperwidth]{\begin{minipage}{21cm}\centering
      \textcolor{gray}{This article has been accepted for publication in the IEEE Internet of Things Journal (IoT-J).\\ DOI: 10.1109/JIOT.2025.3609011} 
    \end{minipage}}}}%
  }
  \AtPageLowerLeft{%
    \raisebox{20pt}{\makebox[\paperwidth]{\begin{minipage}{21cm}\centering
      \textcolor{gray}{ \copyright 2025 IEEE.  Personal use of this material is permitted.  Permission from IEEE must be obtained for all other uses, in any current or future media, including reprinting/republishing this material for advertising or promotional purposes, creating new collective works, for resale or redistribution to servers or lists, or reuse of any copyrighted component of this work in other works.
      }
    \end{minipage}}}%
  }
}
\begin{document}

\renewcommand{\arraystretch}{1.15}

\title{Efficient and Accurate Downfacing Visual Inertial Odometry}

\author{Jonas Kühne,~\IEEEmembership{Graduate Student Member,~IEEE}, 
Christian Vogt,~\IEEEmembership{Member,~IEEE},\\ 
Michele Magno,~\IEEEmembership{Senior Member,~IEEE},
and Luca Benini,~\IEEEmembership{Fellow,~IEEE}
        % <-this % stops a space
\thanks{This work was supported by the Swiss National Science Foundation’s TinyTrainer project under Grant number 207913.}
\thanks{Jonas Kühne is with the Integrated Systems Laboratory and the Center for Project-Based Learning, ETH Zurich, 8092 Zurich, Switzerland (e-mail: kuehnej@ethz.ch).}
\thanks{Christian Vogt is with the Center for Project-Based Learning, ETH Zurich, 8092 Zurich, Switzerland (e-mail: christian.vogt@pbl.ee.ethz.ch).}
\thanks{Michele Magno is with the Center for Project-Based Learning, ETH Zurich, 8092 Zurich, Switzerland (e-mail: michele.magno@pbl.ee.ethz.ch).}
\thanks{Luca Benini is with the Integrated Systems Laboratory, ETH Zurich, 8092 Zurich, Switzerland, and also with the Department of Electrical, Electronic and Information Engineering, University of Bologna, 40136 Bologna, Italy (e-mail: luca.benini@unibo.it).}
\thanks{Copyright \copyright 2025 IEEE. Personal use of this material is permitted. However, permission to use this material for any other purposes must be obtained from the IEEE by sending a request to pubs-permissions@ieee.org.}% <-this % stops a space
}

%\markboth{Manuscript Submitted to \emph{IEEE IoT-J}}%

%\IEEEpubid{0000--0000/00\$00.00~\copyright~2025 IEEE}

\maketitle

\begin{abstract}
Visual Inertial Odometry (VIO) is a widely used computer vision method that determines an agent's movement through a camera and an IMU sensor. This paper presents an efficient and accurate VIO pipeline optimized for applications on micro- and nano-UAVs. The proposed design incorporates state-of-the-art feature detection and tracking methods (SuperPoint, PX4FLOW, ORB), all optimized and quantized for emerging RISC-V-based ultra-low-power parallel systems on chips (SoCs). Furthermore, by employing a rigid body motion model, the pipeline reduces estimation errors and achieves improved accuracy in planar motion scenarios. The pipeline's suitability for real-time VIO is assessed on an ultra-low-power SoC in terms of compute requirements and tracking accuracy after quantization. The pipeline, including the three feature tracking methods, was implemented on the SoC for real-world validation. This design bridges the gap between high-accuracy VIO pipelines that are traditionally run on computationally powerful systems and lightweight implementations suitable for microcontrollers. The optimized pipeline on the GAP9 low-power SoC demonstrates an average reduction in RMSE of up to a factor of 3.65x over the baseline pipeline when using the ORB feature tracker. The analysis of the computational complexity of the feature trackers further shows that PX4FLOW achieves on-par tracking accuracy with ORB at a lower runtime for movement speeds below 24 pixels/frame.

\end{abstract}

\begin{IEEEkeywords}
Constrained Devices, Embedded Devices, Energy Efficient Devices, Cyber-Physical Systems, Mobile and Ubiquitous Systems, Real-Time Systems
\end{IEEEkeywords}

\section{Introduction}

\IEEEPARstart{V}{isual} Inertial Odometry (VIO) describes the process of determining an agent's movement through the use of camera and Inertial Measurement Unit (IMU) data \cite{yao2024sg}. Cameras are used in pure Visual Odometry (VO) to generate a movement estimate from one frame to another by considering the displacement of features or brightness patches between camera images \cite{lin2024deep}. While stereo VO (i.e., using two cameras) can estimate metric depth information through extrinsic calibration, monocular VO can only estimate relative pixel movements. It lacks an absolute scale but shows little drift over time. IMUs, on the other hand, are capable of obtaining metric measurements \cite{lin2024deep} by measuring linear acceleration and rotational velocity. Although the odometry could be estimated purely from IMU data, it is inaccurate due to measurement noise and bias, leading to high estimation errors and, therefore, drift of the odometry signal \cite{wang2022llio}. To compensate for this, VIO utilizes the complementary nature of (monocular) VO and IMU data to produce a motion prediction with little drift and a metric scale \cite{macario2022comprehensive}.

\begin{figure}[t]
    \centering
    \includegraphics[width=\linewidth]{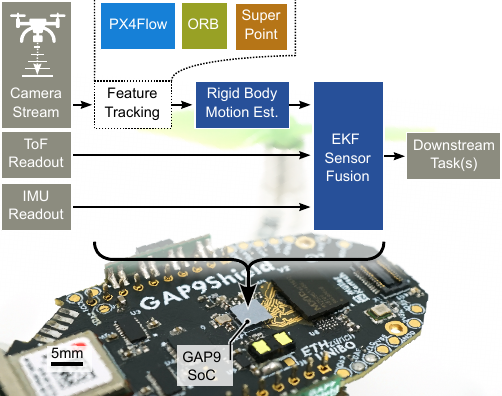}
    \caption{In this work, we present a downfacing VIO pipeline that is suitable for resource-constrained microcontrollers and SoCs used in small-scale UAVs, e.g., the pictured GAP9 shield \cite{muller2024gap9shield}. As feature detectors and trackers, we investigate the classical ORB descriptor \cite{rublee2011orb} and the machine-learned SuperPoint descriptor \cite{detone2018superpoint} and compare both approaches to the existing parallelized PX4FLOW \cite{honegger2013open, kuhne2022parallelizing} implementation.}
    \label{fig:visual_abstract}
\end{figure}

\IEEEpubidadjcol

VIO systems have been well-researched and miniaturized to a certain extent, specifically targeting smartphones \cite{lin2024deep} and mini drones \cite{hanover2024autonomous}. To allow the use of highly accurate VIO in micro- and nano-drones, as well as in AR glasses, these capable systems need to be scaled down further \cite{bahnam2021stereo,shen2023orb}. 

In the literature, we can identify two directions in VIO research. There is work on: 
\begin{enumerate*}[label=(\roman*)]
\item accurate but resource-demanding VIO pipelines that typically run on systems that feature an operating system and can rely on powerful libraries such as \emph{OpenCV} and \emph{Ceres} thanks to the simplified memory handling and abstraction of parallelization \cite{bahnam2021stereo, campos2021orb, kuhne2024low}. And 
\item heavily optimized bare-metal implementations on low-power microprocessors \cite{honegger2013open, kuhne2022parallelizing}. These systems usually rely on much simpler, more lightweight algorithms than those found in the OpenCV library. Despite its age, PX4FLOW \cite{honegger2013open} is a prominent example of the latter category. It is a downfacing\footnote{Downfacing means the camera is oriented towards the ground, i.e., facing in the direction of the gravity vector.} VIO system often used in drone applications, which was published over a decade ago.
\end{enumerate*}

The main contribution of this paper is to bridge the gap between the two above-mentioned directions. With the emergence of ultra-low power parallel Systems on Chips (SoCs), including those based on the RISC-V architecture, more computational resources, and efficient processing are available within the power budget of classic micro-controller units (MCUs), such as those used for  PX4FLOW \cite{muller2024gap9shield}. These new platforms could enable VIO to achieve performance comparable to higher-performance processors (i.e., Smartphones, ARM Application Processors) within small, low-power systems. However, achieving such efficiency comes with several challenges, including designing highly efficient models that exploit techniques such as quantization and parallelization \cite{saha2022machine}. In particular, VIO pipelines for microcontrollers and low-power SoCs require careful balancing of computational load and latency and ensuring precision, all while staying within strict power and memory constraints \cite{he2021picovo,saha2022machine}.

We propose a VIO pipeline that leverages the computational advancements of ultra-low-power SoCs by implementing quantized variants of ORB \cite{rublee2011orb} and SuperPoint \cite{detone2018superpoint}, showing real-time performance on the SoC hardware at low power. We present an efficient (in terms of operations) and accurate downfacing VIO pipeline optimized for micro- and nano-UAV applications. We focus on the UAV application of downfacing VIO akin to PX4FLOW, essentially reducing the estimation problem from six degrees of freedom to only four degrees of freedom. In addition to PX4FLOW, which uses lightweight image-patch-based optical flow estimation, we investigate the suitability of the two state-of-the-art feature-matching methods ORB \cite{rublee2011orb} and SuperPoint \cite{detone2018superpoint} for low-power and real-time VIO. Furthermore, we propose the use of rigid body motion assumption to decompose the flow in x, y, and yaw components \cite{tenenbaum2006fundamentals, bergen1992hierarchical} to more accurately account for rotations instead of using averaged flow values in the x and y directions. As the rigid body motion estimation suffers from outliers, we additionally present a lightweight outlier rejection scheme.

To benchmark our pipeline variants, we compare the VIO accuracy and the computational requirements to the baseline implementation of the parallelized PX4FLOW presented in \cite{kuhne2022parallelizing}. We benchmark the classical ORB descriptor, the machine-learned SuperPoint \cite{detone2018superpoint} descriptor and a modified PX4FLOW variant against this baseline implementation. Lastly, we deploy those three pipeline variants on an ultra-low-power multi-core GAP9 SoC\footnote{GreenWaves GAP9: \url{https://greenwaves-technologies.com/gap9_processor/}}, which has been successfully deployed on nano-drones\cite{muller2024gap9shield} and investigate the resulting computational load and latency in comparison to the parallelized PX4FLOW derivative \cite{kuhne2022parallelizing}.

In summary, this paper aims to improve downfacing VIO while leveraging the additional computational resources that novel ultra-low-power parallel SoCs provide. The contributions are the following:

\begin{itemize}
    \item We investigate the viability of various feature detectors and trackers (ORB \cite{rublee2011orb}, SuperPoint \cite{detone2018superpoint}, and PX4FLOW \cite{kuhne2022parallelizing}) for resource-constrained downfacing VIO in terms of accuracy and latency on a common estimation pipeline.
    \item We improve the VIO prediction accuracy by estimating rigid-body motion on the tracked feature displacement (while rejecting displacement outliers) instead of a weighted average calculation.
    \item For evaluating our approach, we present a full-fledged downfacing VIO pipeline completely implemented and tested on the low-power SoC GAP9.
    \item To ensure the continued development of VIO on resource-constrained devices, we open-source our GAP9 implementations of the feature trackers and the complete VIO pipeline. In addition, we also provide a hardware-agnostic ORB variant implemented using integer representation. \\ \url{https://github.com/ETH-PBL/Downfacing-VIO}
\end{itemize}

The remainder of this article is organized as follows. \Cref{sec:related_work} presents related work in a top-down fashion. Furthermore, we explain the used taxonomy and categorize our VIO approach. In \Cref{sec:methods}, we introduce the template VIO pipeline, which is used in conjunction with the three feature trackers. Furthermore, we detail how we ported the ORB and SuperPoint feature trackers to GAP9. In \Cref{sec:evaluation}, we elaborate on the hardware setup and dataset used for the evaluation of the various pipeline variants. We provide and assess the experimental results in \Cref{sec:results}. In \cref{sec:discussion}, we discuss the performance of the proposed VIO system and give recommendations for the use of our approach in real-world applications. \Cref{sec:conclusion} concludes this article.

\section{Related Work}
\label{sec:related_work}

VIO and Visual Inertial SLAM pipelines are widely investigated in the visual perception field \cite{macario2022comprehensive}. Therefore, we approach the related work in a top-down fashion. Starting with a broader overview of the topic of VIO and subsequently narrowing the scope to down-facing implementations of VIO and resource-constrained computing platforms.

\subsection{Visual Inertial Odometry Overview and Terminology}
\label{sec:related_work_terminology}

VIO systems estimate the movement of an agent carrying a camera and an IMU in a scene, tracking a full six degrees of freedom movement. VIO systems can roughly be categorized according to the following dimensions:

\begin{itemize}
    \item \textbf{Direct versus feature-based methods:} Direct methods estimate the motion between two frames by optimizing the reprojection error of image patches with respect to the motion parameters \cite{von2018direct}. In contrast, feature-based (indirect) methods estimate the motion through feature detection, description, and matching, where image points get assigned a descriptor and are triangulated into a world coordinate system. Using the descriptors of image features in a new frame, the relative position of the camera to the world can be estimated using the previously triangulated features by solving the \emph{Perspective-n-Point} problem \cite{mourikis2007multi,leutenegger2014keyframe,qin2018vins}. Furthermore, some algorithms use a combination of both concepts (hybrid) \cite{bloesch2015robust} or take a layered approach (semi-direct) and perform a direct method on a frame-to-frame basis in combination with a feature-based method on distinct frames typically called keyframes.
    \item \textbf{Filtering versus optimization-based methods:} Once the relative movement between two frames has been obtained, this estimate can be refined. An efficient way to do so is by using filtering methods like Extended Klaman Filters (EKFs) \cite{mourikis2007multi,bloesch2015robust}. More resource-demanding, but also more accurate are optimization methods, where the movement estimate between potentially multiple earlier frames is refined using the reprojection error. With these methods, multiple parameters can be optimized like the movement estimates, the triangulated world coordinates of the features, and potentially also the camera and IMU parameters \cite{leutenegger2014keyframe,qin2018vins,campos2021orb,von2018direct}.
    \item \textbf{Loop closure:} Another dimension to distinguish VIO pipelines is whether they apply loop closure. Loop closure is the process of detecting previously visited places to correct for drift and accumulation errors. Sometimes, the terms Odometry and SLAM are used to indicate systems without and with loop closure respectively \cite{campos2021orb,qin2018vins}.
\end{itemize}

\Cref{tab:rw_vio_algorithms} gives an overview and categorization of the most influential VIO and Visual Inertial SLAM papers surveyed in greater detail in \cite{macario2022comprehensive}.

The approaches presented in this work are all feature-based methods. The main differentiation lies in the used features, ranging from raw image patches over the binary ORB descriptor \cite{rublee2011orb} to the machine-learned SuperPoint descriptor \cite{detone2018superpoint}. All approaches share a common Kalman Filter architecture for the refinement of the obtained poses, and no loop closure is applied.

\begin{table}[t]
    \centering
    \caption{Influential related work in monocular visual-inertial odometry as presented in \cite{macario2022comprehensive}.}
    \aboverulesep=0ex
    \belowrulesep=0ex
    \begin{tabular}{l|l|l|l|l}
        \toprule
        \textbf{Algorithm} & \textbf{Year} & \textbf{Tracking} & \textbf{Refinement} & \textbf{LC} \\
        \midrule
        \textbf{MSCKF \cite{mourikis2007multi}} & 2007 & Feature-based & Filtering & No \\
        \textbf{OKVIS \cite{leutenegger2014keyframe}} & 2014 & Feature-based & Optimization & No \\
        \textbf{ROVIO \cite{bloesch2015robust}} & 2015 & Hybrid & Filtering & No \\
        \textbf{VINS-Mono \cite{qin2018vins}} & 2018 & Feature-based & Optimization & Yes \\
        \textbf{VI-DSO \cite{von2018direct}} & 2018 & Direct & Optimization & No \\
        \textbf{ORB-SLAM3 \cite{campos2021orb}} & 2021 & Semi-direct & Optimization & Yes \\
        \bottomrule
    \end{tabular}
    \label{tab:rw_vio_algorithms}
\end{table}

\subsection{Downfacing Visual (Inertial) Odometry}
A simplification over the full six degrees of freedom motion estimation is the restriction to planar motions, only estimating translations parallel to the ground as well as rotations around the yaw axis. While this already allows for motion estimation along three degrees of freedom, these systems sometimes additionally estimate the height of the agent for four degrees of freedom.

PX4FLOW successfully applied this idea to enable full-onboard processing of downfacing VO (no IMU was used) on an STM32F407 microcontroller using a camera, a gyroscope, and an ultrasonic distance sensor \cite{honegger2013open}. The original PX4FLOW implementation is restricted to a movement of $\pm$4 pixels per 64-by-64 pixel frame for the selected imaging sensor, lens, and frame rate of 250 FPS. This results in a trackable velocity of $\pm$1.5 meters per second for a distance of one meter to the ground \cite{honegger2013open}.

A follow-up work on PX4FLOW showed that a parallelized and improved version of the PX4FLOW algorithm can reach significantly higher frame rates (more than 500 FPS) on a more recent System on Chip called GAP8 while staying within the power envelope of the original implementation \cite{kuhne2022parallelizing}.

Utilizing a larger setup, the authors of \cite{shen2023multi} present a solution based on a 3x3 camera array with varying apertures that produce more robust optical flow estimates in low-altitude flights than single-camera systems. Additionally, through extrinsic calibration between the nine cameras, the multi-aperture camera array can estimate the distance to the ground.

The very high frame rates of $>$ 250 FPS needed by PX4FLOW for tracking $\pm$1.5 meters per second are at the limit of what lightweight small-scale cameras can provide \cite{gove2020cmos}. Therefore, this paper considers more recent feature-based methods that can track much larger (arbitrarily large) pixel displacements than PX4FLOW derivatives, requiring lower frame rates to track similar or even higher movement speeds than PX4FLOW. 

\subsection{VIO on Resource-Constrained Platforms}

V(I)O has been implemented on various resource-constrained devices. Starting with devices that still offer significant computational resources like smartphones \cite{lin2024deep} and Raspberry Pis \cite{bahnam2021stereo,kuhne2024low} and moving down to microcontroller devices and SoCs \cite{honegger2013open,kuhne2022parallelizing,he2021picovo}. For both MSCKF \cite{mourikis2007multi} and VINS-Mono \cite{qin2018vins}, modified versions have been implemented on Raspberry Pis. The lightweight S-MSCKF \cite{sun2018robust} algorithm has been implemented on a Raspberry Pi Zero as presented in \cite{bahnam2021stereo}. In contrast, for VINS-Mono, a more powerful Raspberry Pi Compute Module 4 plus additional on-sensor acceleration for the computation of Optical Flow was used as presented in \cite{kuhne2024low}. Microcontroller implementations can be found in the already mentioned PX4FLOW variants \cite{honegger2013open,kuhne2022parallelizing}. Additionally, PicoVO \cite{he2021picovo} presents a lightweight implementation of a six-degree-of-freedom VO system running on an STM32F767 microcontroller at 33 FPS on average processing images at QVGA (320x240) resolution.

In this work, we use a 10-core GAP9 SoC to implement a downfacing VIO system, showcasing three feature extraction methods. We operate on QQVGA (160x120) inputs to compare our different methods. Using a VICON motion capture system, we quantify the tracking accuracy improvements over the PX4FLOW baseline. We run the different feature-tracking methods in combination with our proposed VIO pipeline that leverages the computational resources available on the GAP9 low-power SoC to improve tracking accuracy.

\section{Methods}
\label{sec:methods}
In this section, we discuss the proposed algorithm and the software pipeline designed for downfacing VIO estimation. For the proposed VIO approaches to be suitable for low-power devices, we avoid costly operations: Following the taxonomy presented in \Cref{sec:related_work_terminology}, we use feature-based tracking methods in combination with a filtering approach for the fusion of inertial and visual estimates. Furthermore, we omit any non-linear optimizations, such as bundle adjustment and loop closure. We detail the optimized and quantized template pipeline, which provides a common base infrastructure into which we plug in the three feature trackers for fair comparison. Furthermore, we elaborate on the measures that were taken to deploy each of the feature trackers on GAP9 while adhering to the strict memory constraints. Since the compute cluster cores of GAP9 share four floating-point units (up to 3.3 GFLOPS for 32-bit data), we use low-precision fixed-point arithmetic where possible (up to 15.6 GOPS for 8-bit data) \cite{valente2024heterogeneous}. For the template pipeline and the feature trackers, we indicate how we leverage the parallel processing capabilities of GAP9.

\subsection{Template VIO Pipeline}\label{sec:template_vio_pipeline}
\begin{figure}[t]
    \centering
    \includegraphics{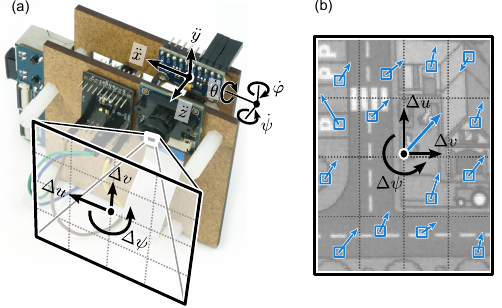}
    \caption{Coordinate systems of the VO and IMU relative to the data collection setup (a) and overview of the rigid body motion estimation with features on the camera image (boxes, small arrows) and estimated overall movement direction in $\Delta u$, $\Delta v$, and $\Delta \psi$ (b).}
    \label{fig:hw_architecture2}
\end{figure}

To ensure a fair and consistent comparison of the three feature tracking methods, the remainder of the software pipeline remains identical, allowing us to isolate and evaluate the performance of each tracker independently. The template VIO pipeline consists of the following stages, as depicted in \Cref{fig:visual_abstract}:

\begin{enumerate}
    \item \textbf{Sensor Readout:} The three sensors (camera, IMU, and ToF sensor) are read out by the fabric controller core. The camera and IMU are read out at the same rate, which depends on the processing time required by the selected feature tracker. The ToF sensor, which mainly provides reference height measurements, is configured in its max range mode, allowing a read-out rate of 6.94\,Hz. Once the sensor data is available, it is transferred to the L1 memory of GAP9, such that it is available to the cluster cores. All the subsequent processing is then performed on the cluster.
    \item \textbf{Feature Tracking:} In this stage, we plug in the various feature trackers, i.e., parallelized PX4FLOW \cite{kuhne2022parallelizing}, ORB \cite{rublee2011orb}, or SuperPoint \cite{detone2018superpoint}, as described in the subsequent sections. From the feature trackers, we obtain optical flow predictions (i.e., the displacement of each feature from one frame to the next).
    \item \textbf{Rigid Body Decomposition:} We model the movement as a rigid body movement \cite{tenenbaum2006fundamentals} and set the origin of the coordinate system to the center of the camera image. Using this assumption, we can decompose the movement into the translational parts $\Delta u$ and $\Delta v$, denoting movement in pixels along the x- and y-direction, respectively, and the yaw-rotation $\Delta \psi$ (\Cref{fig:hw_architecture2}). To account for outliers in the optical flow prediction, we implement the following outlier rejection:

    \begin{enumerate}
    \item In the first step, we build a histogram of the movement magnitudes in the $x$ and $y$ directions. Per direction, we select the bin with the most entries as the baseline. We then consider those optical flow predictions as inliers, which are within five pixels of the baseline prediction. We solve the equation using all the inlier points to obtain a preliminary rigid body motion estimate.

    \item Using the previous result, we again classify the optical flow predictions as in- and outliers by applying the obtained motion estimate to the features. If the estimated feature position from the motion estimate is within 1.5 pixels of the position of the tracked feature, it is considered an inlier. Using the inliers, the equation is solved a second time.
    \end{enumerate}

    \item \textbf{Kalman Filtering with the IMU states:} We use an Extended Kalman Filter \cite{ribeiro2004kalman} to fuse the states of the rigid body decomposition (i.e., $\Delta u$, $\Delta v$, and $\Delta \psi$) with the states of the IMU (i.e., $\ddot{x}$, $\ddot{y}$, $\ddot{z}$, $\dot{\varphi}$, $\dot{\theta}$, and $\dot{\psi}$). We use $\Delta$ to denote changes between two frames and the derivative notation to denote derivatives in time. 
    %\todo{explain how z can be estimated}
    \item \textbf{Providing the filtered states to downstream tasks:} After the Kalman filtering, the lateral and rotational states are available to downstream tasks. Depending on the application, those can be absolute positions and orientations or acceleration and velocity information. For our validation, we use absolute positions and orientations.
\end{enumerate}

In addition to the template pipeline, we implement the pipeline presented in PX4FLOW \cite{honegger2013open} as a reference model. The reference model does not perform a rigid body decomposition and only has access to the gyroscope states of the IMU, i.e., $\dot{\varphi}$, $\dot{\theta}$, and $\dot{\psi}$.

%\content{Explain the basic structure of the pipelines (i.e. flow + IMU are fused in a KF), use rigid-body assumption?}

\subsection{ORB}
For this work, the original ORB implementation presented in \cite{rublee2011orb}, which is part of the \emph{OpenCV} library, has been ported from \emph{C++} to \emph{C} and quantized to low-precision integer representations where possible. The modifications to the ORB pipeline stages are described below:

\begin{enumerate}
    \item \textbf{FAST Corner Detection:} The original ORB implementation uses the \emph{FAST} algorithm \cite{rosten2006machine} to detect corners. Since the algorithm compares the pixel values of 16-pixel locations around a center pixel with the value of the center pixel and additionally derives a \emph{FAST-Score} by summing the absolute differences, the algorithm could be implemented using 8-bit and 16-bit integer values without any loss of accuracy over the original algorithm.
    \item \textbf{Harris Corner Detection:} The \emph{Harris} corner detection implementation \cite{harris1988combined} is inspired by the \emph{OpenCV} implementation. For the calculation of the image gradients $I_x$ and $I_y$ in $x$ and $y$ direction, respectively, we use the same patch-size of 7-by-7 and a first order \emph{Sobel}-filter with kernel-size three as is used in \emph{OpenCV}. With those configurations, we are able to represent the entries of the $M$ matrix as defined in \cite{harris1988combined} with 32-bit integers without any loss of accuracy. Before the calculation of the \emph{Harris}-score $R$, the entries of the matrix $M$ are quantized to 16-bit integers by scaling with a factor of $2^{-11}$ such that the resulting value of $R$ can be represented as a 32-bit signed integer. The \emph{Harris}-score is defined as
    \begin{equation}
        R = det(M) - k(trace(M))^2,
    \end{equation}
    where $det(M)$ and $trace(M)$ are the determinant and trace of the matrix $M$ respectively and $k$ is a design parameter, typically set to $0.04$ \cite{sanchez2018analysis}. Note that different from the \emph{Harris} implementation in \emph{OpenCV} where the score is represented as a floating-point number between zero and one and therefore is normalized by the size of the patch, the \emph{Sobel}-filter entries, and the pixel value-range (i.e., $1/(7 \cdot 4 \cdot 255)^4$), our score is scaled by $(2^{-11})^2$ and represented as a signed 32-bit integer. Since the \emph{Harris}-score is used to reject features with a low score based on a threshold value and as a relative measure to sort feature candidates by the corner value (i.e., the score), we can account for this different scaling by adjusting the threshold value accordingly. Furthermore, the quantization of the entries of the $M$ matrix only impacts the accuracy of $R$ values close to zero. Since the threshold to select features is significantly larger than zero ($R_{threshold} \gg 0$), the quantization does not impact the accuracy of the corner detection.
    \item \textbf{Image Blurring:} The image blurring is slightly simplified in comparison to the \emph{OpenCV} implementation. For numeric stability and simplicity, we use an approximated 5-by-5 \emph{Gaussian} filter kernel represented in 8-bit unsigned integer values. The values of the filter kernel sum to 256, guaranteeing that the accumulated filter value can be represented as a 16-bit unsigned integer before being normalized by 256. For the 2-pixel boundary of the image, we do not apply any padding and copy the original pixel-value into the filtered image. In comparison, in the \emph{OpenCV} implementation, a floating point 7-by-7 \emph{Gaussian} filter kernel is used, and the image boarders are reflected (i.e., mirrored) as a padding strategy.
    \item \textbf{Feature De-Rotation and Description:} The ORB algorithm \cite{rublee2011orb} determines a dominant orientation for every feature by calculating the intensity-weighted centroid in a circular image patch around a previously detected feature location. The bit pattern used to describe the features is rotated according to this dominant orientation before determining the feature descriptor. Since we are only interested in the direction of the centroid and not in its position, we can omit scaling the accumulated value by the number of pixels in the patch. Therefore, we can sum the intensity weighted offsets from the center of the pixel patch using a 32-bit signed integer for both the x- and y-offsets. We use a floating point implementation of the arctan function that preserves the information about the quadrant (i.e., \texttt{atan2}) to determine the angle of the dominant orientation. To determine the feature descriptors, we use the original bit-pattern of \cite{rublee2011orb} and rotate it according to the dominant direction. To rotate the bit-pattern, we calculate the sine and cosine values of the rotation angle using the respective floating point implementation, and represent the values as signed Q7.8 fixed-point numbers when calculating the rotated bit-pattern. After rotating the bit pattern, we round the values to the nearest integer before obtaining the binary feature descriptor.
    \item \textbf{Feature Matching:} The feature matching uses a brute-force approach. For every feature in frame $n$, the best match in frame $n-1$ is determined. The similarity of the two features is computed as the hamming distance between both binary feature descriptors; the smaller the hamming distance, the more similar the two features are. We consider two features a match once the hamming distance is 20 or lower (for feature descriptors of length 256).
\end{enumerate}

The full ORB pipeline can be executed on a single core or in a parallelized fashion across GAP9's eight worker cores.

\subsection{SuperPoint}
We use the SuperPoint algorithm described in \cite{detone2018superpoint} to compare with a recent machine-learned feature tracker with a small model size. We use the implementation and checkpoint provided by \emph{Magic Leap}\footnote{\url{https://github.com/magicleap/SuperPointPretrainedNetwork}}, the authors of the SuperPoint paper \cite{detone2018superpoint}. We use the \emph{NNTool} utility of the Software Development Kit (SDK) provided by \emph{GreenWaves Technologies} to quantize and deploy the SuperPoint model on GAP9. Using the \emph{ONNX} file of the SuperPoint model, plus a representative set of sample inputs, the \emph{NNTool} determines an optimal weight and activation quantization. Additionally, the \emph{NNTool} matches the network operations to the available compute resources. To fit the SuperPoint network onto GAP9, we use 8-bit quantization for activations and weights.

In the case of SuperPoint, predictions about feature locations and descriptors are made on a 20x15 grid (i.e., the output is subsampled by a factor of 8x in both x- and y-directions). The feature locations are encoded as likelihoods in a 64x20x15 output that can be converted to the original image size, and the descriptors get upsampled to the original 160x120 image input through interpolation at the corresponding pixel location. The features between two frames are matched using a similar brute force matcher implementation as for the ORB descriptors. SuperPoint also produces a descriptor of length 256, but in contrast to ORB, the descriptor entries are 8-bit integers (for the quantized model). Instead of using the hamming distance, we use the cosine similarity to measure the similarity of two descriptors.

\subsection{Baseline Implementation: Parallelized PX4FLOW}
For the parallelized PX4FLOW variant, we use the implementation described in \cite{kuhne2022parallelizing} and port it to GAP9. As the original algorithm is already implemented in fixed-point logic, no further optimizations or quantization are needed. Since the original pipeline only returns the flow vectors but not the absolute coordinate in the image frame, we added this information to the output values to be able to apply our rigid body motion estimation. 

Since the parallelized PX4FLOW variant functions as a baseline, we did not make any functional changes to its feature-tracking approach. However, we did omit the magnetometer in the IMU measurements. For comparison purposes, we present the performance of both PX4FLOW in the original configuration (i.e., without rigid body motions estimation) \cite{honegger2013open,kuhne2022parallelizing} and when integrated into our template pipeline described in \Cref{sec:template_vio_pipeline}.

\section{Evaluation Methodology}
\label{sec:evaluation}

To evaluate the computational load of the pipeline configurations, we analyzed the cycle count and the end-to-end latency of the proposed methods on GAP9. To assess the accuracy of the pipelines, we built a hardware platform. We recorded several benchmarking sequences, including the necessary sensor modalities and ground truth position data, as described in the following sections.

\subsection{Hardware Platform}
\begin{figure}[ht]
    \centering
    \includegraphics{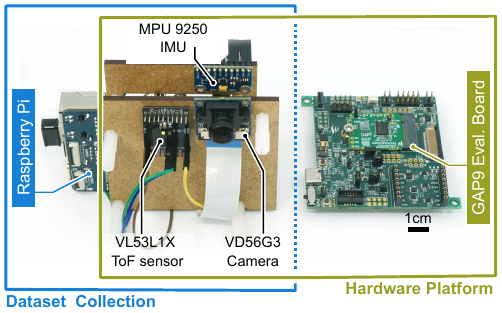}
    \caption{Physical setup for data collection, with the camera, IMU, and ToF sensor wired to a Raspberry Pi, as well as the benchmarking and evaluation platform where the Raspberry Pi is replaced by the GAP9 Microcontroller.}
    \label{fig:hw_architecture}
\end{figure}

The hardware platform shown in \Cref{fig:hw_architecture} is based on the GAP9 SoC by GreenWaves Technologies. GAP9 features 10 RISC-V cores, of which nine cores constitute a compute cluster, with a master and eight worker cores, plus an additional RISC-V core termed fabric controller, that orchestrates the interaction with all peripherals. Both the compute cluster and fabric controller support a clock frequency of up to 370\,MHz. Furthermore, GAP9 provides 1.5\,MB interleaved L2 memory and 128\,KB L1 memory that is shared among the 9 compute cluster cores.

As a camera sensor, we use the VD56G3 global shutter sensor by STMicroelectronics, supporting a resolution of up to 1124x1364 (at 88 FPS) or VGA (640x480 at 237 FPS)\footnote{\url{https://www.st.com/en/imaging-and-photonics-solutions/vd56g3.html}}. Since we are performing our comparison using QQVGA resolution, the sensor can be configured to even higher frame rates than 237 FPS. We assume single-lane MIPI CSI-2 communication at 1.5\,Gbit/s as a limiting factor, dictating the frame rate to image resolution trade-off. Specifically, for the evaluation of the parallelized PX4FLOW variant of \cite{kuhne2022parallelizing}, we use a frame rate of 300 FPS. The VD56G3 sensor is connected via a single-lane MIPI CSI-2 interface to the GAP9 SoC.

We chose the MPU 9250 by TDK InvenSense as an IMU and connected it to the GAP9 using the IMU's I2C interface. Lastly, we substitute the ultrasonic distance sensor of PX4FLOW with a more lightweight Time-of-Flight (ToF) sensor. We deploy a downfacing VL53L1X ToF sensor by STMicroelectronics to provide reference distance measurements to the floor. The sensor can be configured for fast or accurate operation. Since we are interested in an accurate reference measurement, we operate the sensor in its accurate setting, achieving a repeatability error as low as $\pm0.15\%$.

\subsection{Dataset}
For evaluating our proposed pipeline, we recorded a benchmarking dataset containing indoor and outdoor sequences.

\begin{figure}[th]
    \centering
    \includegraphics[width=\linewidth]{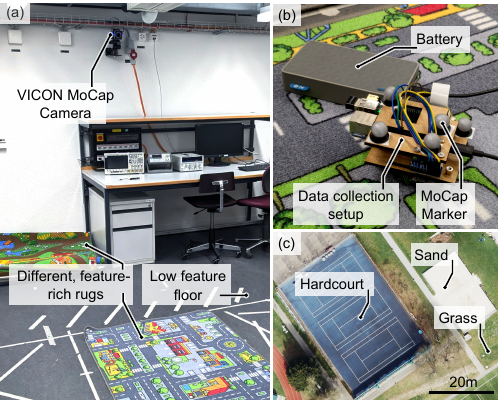}
    \caption{Data collection setup overview with different floor types and motion capture (MoCap) system (a), as well as data collection setup (\Cref{fig:hw_architecture}) with MoCap markers (b), and aerial photograph of the outdoor test area (c), adapted from \cite{orthofoto_zh}.}
    \label{fig:data_collection_setup}
\end{figure}

The indoor dataset contains seven movement trajectories of the sensor system depicted in \Cref{fig:data_collection_setup}b. In addition to the sensor data, the dataset contains the ground truth poses of the system captured by a Vicon motion capture system. The movement sequences were recorded in a space of 4-by-4 meters, performing various movement patterns and using two different floor textures. The first texture consisted of the rugs as displayed in \Cref{fig:data_collection_setup}a, which provides rich, distinctive features, and the second texture was the uniformly colored floor of the room, which provides very few features to track.

The outdoor dataset was recorded in an outdoor sports facility, offering larger space and various surfaces to test a VIO pipeline. We recorded sequences on hardcourt, sand, and grass, as well as a sequence combining the three surfaces, plus parts of a cobblestone path. The ground truth poses were captured with GPS-RTK.

The indoor dataset allows the evaluation of the VIO performance under specific movement patterns like pure translation, movement in a square, or random movement, whereas the outdoor dataset provides real-world conditions and varying surface textures.

\subsection{Latency and Cycle Count Profiling}
To analyze the latency and cycle count of the various pipelines, we executed them on GAP9 and used the performance counters available on the hardware. Additionally, we used the GAP Virtual SoC (GVSoC) simulation tool to simulate the execution on the GAP9 SoC.

We indicate the cycle counts of the various sub-steps within the feature trackers. For the trackers, where we have single- and multi-core implementation, we additionally measure the achieved parallelization speed-up.

\subsection{Accuracy of Feature Trackers and VIO Pipeline}
For the accuracy measurements, we ran the various pipelines at a hypothetical 100 FPS on the GVSoC. The accuracy of the trackers and pipelines is measured using the benchmarking dataset containing indoor and outdoor sequences. We measure both the RMSE and relative translation error with respect to either the motion capture (indoor) or GPS-RTK (outdoor) ground truth. We indicate those two metrics since an early error in the orientation prediction leads to a high RMSE, even if the subsequent tracking is again accurate. The relative translation error provides a metric that more holistically represents tracking accuracy. To calculate and evaluate those two error metrics, we use the \emph{RPG Trajectory Evaluation}\footnote{\url{https://github.com/uzh-rpg/rpg_trajectory_evaluation}} toolkit \cite{zhang2018tutorial}. For the calculation of the RMSE, we use the first 10 seconds of the prediction (i.e., 1000 frames) to align the VIO prediction with the indoor motion capture ground truth using the sim(3) transformation \cite{zhang2018tutorial}. In the outdoor experiments, we use 30 seconds of the prediction (i.e., 300 frames), due to the lower 10\,Hz frequency of the GPS-RTK ground truth system. For the relative translation error, the sub-trajectory start point of the VIO prediction is set to the ground truth value and diverges from there (i.e., corresponding to an alignment over one frame or pose).

\section{Results}
\label{sec:results}

This section presents the experimental results of validating the various pipeline configurations on the GAP9 SoC. We profile the computational effort and discuss the VIO accuracy of the various configurations.

\subsection{Profiling}
We analyzed the cycle count and latency of the various pipelines when executing on GAP9 in \Cref{tab:cycles_counts}. For a fair comparison, we used the same compiler optimization level for all algorithms. Since we are mainly interested in a speedy execution, we went for the highest optimization level, i.e., \texttt{-O3}. Additionally, we operate all the pipelines with a clock speed of 370\,MHz for minimal latency. Studies have shown that GAP9 uses between 64\,mW and 68\,mW at 370\,MHz when utilizing the nine-core compute cluster \cite{kiamarzi2024qr, cereda2024device}. For the pipeline steps that can be run both parallelized and in single core mode, we additionally indicate the parallelization speed up. 

It is worth mentioning that the reception of new images and the transfer from L2 to L1 memory is handled using DMA transfers, which the Fabric Controller orchestrates. Although those transfers increase latency, they do not require any processing resources and are fast enough not to impact the maximum achievable frame rate.

\subsubsection{Template Pipeline}
The template pipeline includes the rigid-body motion estimation with outlier rejection and Kalman filtering of the movement estimates and IMU readings. Those two steps are mainly computed sequentially and, therefore, not parallelized, so we do not indicate any parallelization speed-up. The rigid body motion estimation performs a first outlier rejection step as described \cref{sec:template_vio_pipeline}, which is done in integer logic and only then performs two iterative rigid body motion estimations, which are singular-value decompositions on a 2x2 matrices. Due to the necessary floating point operations, the rigid body motion estimation requires 66.9 kcycles. The Kalman Filtering is implemented entirely in floating point. In addition to the filtering itself, the various sensor readings are converted to a common coordinate system (the one of the camera). Since the coordinate transformations involve multiple 4x4 floating point multiplications, the Kalman Filtering module requires 156 cycles.

\begin{table}[t]
    \centering
    \caption{The computational demand of the various feature tracking methods: ORB, Optimized PX4FLOW, and SuperPoint (SP). Additionally, we indicate the computational load of the template pipeline steps, i.e., Rigid Body Estimation and Kalman Filtering.}
    \aboverulesep=0ex
    \belowrulesep=0ex
    \begin{tabular}{l|r|r|r}
        \toprule
        \textbf{Pipeline Configrations} & \textbf{Single-Core} & \textbf{Multi-Core} & \textbf{Speed Up} \\
         & [kCycles] $\downarrow$ & [kCycles] $\downarrow$ & [Factor] $\uparrow$ \\
        \midrule
        \midrule
        ORB - Feature Detection & 7 718.8 & 1 143.6 & 6.75x \\
        ORB - Blurring & 1 785.5 & 245.8 & 7.26x \\
        ORB - Feature Extraction & 7 050.6 & 908.1 & 7.76x \\
        ORB - Matching & 4 688.2 & 601.8 & 7.79x \\
        \midrule
        \textbf{ORB - Total} & 21 243.1 & 2 899.3 & \textbf{7.33x} \\
        \midrule
        \midrule
        \textbf{Optimized PX4FLOW} & 593.1 & \textbf{81.3} & 7.30x \\
        \midrule
        \midrule
        SP - Feature Extraction & - & 30 036.3 & - \\
        SP - Post Processing & 1 577.4 & 206.2 & 7.65x \\
        SP - Matching & 5 496.3 & 732.9 & 7.50x \\
        \midrule
        \textbf{SP - Total} & - & 30 975.4 & - \\
        \midrule 
        \midrule
        \textbf{Rigid Body Estimation} & 66.9 & - & - \\
        \textbf{Kalman Filtering} & 156.0 & - & - \\
        \bottomrule
    \end{tabular}
    \label{tab:cycles_counts}
\end{table}

\subsubsection{Feature Trackers} 
\label{sec:results_profiling_feature_trackers}
Since the feature detectors work directly on the recorded image, which can be split into chunks of similar size, all three feature detectors can be parallelized. 

For the feature detection stage of ORB, we distribute the image across the eight worker cores. The FAST \cite{rosten2006machine} detector is run on every pixel. Depending on how many features are selected by FAST per region, the computational load incurred by the Harris score calculation can vary per core, leading to varying parallelization speed-ups (6.75x on average for our dataset and a target value of 300 descriptors). For the remaining stages of ORB, we distribute the calculations as evenly as possible over all the cores. If the input is divisible by eight, this can lead to a very high parallelization speed-up (i.e., close to 8x) as shown in \Cref{tab:cycles_counts}.

For PX4FLOW, we used the \emph{locally optmized} PX4FLOW implementation of \cite{kuhne2024low}. Since GAP9 is the successor of GAP8 with architectural improvements, we were able to achieve smaller cycle counts (593.1 kcyles vs. 676.1 kcycles in single-core configuration and 81.3 kcycles vs. 88.8 kcycles in multi-core configuration) and a higher parallelization speed up (7.30x vs. 7.21x) than reported in the original paper \cite{kuhne2022parallelizing}.

\begin{table*}[t]
    \centering
    \caption{The table shows the details of each of the recorded indoor benchmarking sequences like movement pattern, used texture, the recording duration, as well as the ground truth length of the recorded trajectory. Furthermore, the benchmarked pipeline configurations, as well as the resulting RMSE in meters, are given.}
    \aboverulesep=0ex
    \belowrulesep=0ex
    \begin{tabular}{l|l|l|c|c|c|c|c|c}
        \cmidrule[\heavyrulewidth]{6-9}
        \multicolumn{5}{c}{} & \textbf{ORB} & \textbf{SuperPoint} & \textbf{PX4FLOW} & \textbf{PX4FLOW} \\
        \cmidrule{5-9}
        \multicolumn{4}{c}{} & Pipeline & Ours & Ours & Ours & Orig. PX4FLOW \\
        \multicolumn{4}{c}{} & Framerate & 100 FPS & 100 FPS & 100 FPS & 300 FPS \\
        \multicolumn{4}{c}{} & Max Movement & $\pm 32$ pixel & Arbitrary & $\pm 4.5$ pixel & $\pm 4.5$ pixel \\
        \cmidrule{5-9}\morecmidrules\midrule
        \textbf{Sequence} & \textbf{Movement} & \textbf{Texture} & \textbf{Duration [s]} & \textbf{ Length [m]} & \multicolumn{4}{c}{\textbf{RMSE [m] $\downarrow$} and (Standard Deviation)} \\
        \midrule
        02 & Square & Rug (rich) & 54.3 & 27.33 & 0.292 (0.149) & 1.860 (0.956) & \textbf{0.275} (0.145) & 4.944 (2.524) \\
        03 & Random & Rug (rich) & 55.3 & 33.54 & 0.499 (0.277) & 2.105 (1.254) & \textbf{0.394} (0.216) & 5.376 (3.450) \\
        04 & Random & Rug (rich) & 43.7 & 37.34 & \textbf{0.348} (0.191) & 3.411 (2.136) & 0.464 (0.262) & 2.254 (1.280) \\
        05 & Translation & Rug (rich) & 49.7 & 25.31 & 0.369 (0.221) & 3.103 (1.675) & 0.320 (0.192) & \textbf{0.140} (0.080) \\
        06 & Square & Floor (sparse) & 53.7 & 32.24 & \textbf{1.613} (1.048) & 3.168 (1.828) & 1.631 (0.984) & 3.941 (1.963) \\
        07 & Translation & Floor (sparse) & 50.0 & 30.75 & 1.544 (0.842) & 5.793 (3.390) & 0.554 (0.296) & \textbf{0.446} (0.215) \\
        08 & Random & Floor (sparse) & 55.7 & 35.45 & \textbf{0.598} (0.255) & 3.287 (2.058) & 1.130 (0.574) & 2.102 (1.063) \\
        \bottomrule
    \end{tabular}
    \label{tab:accuracy_results_rmse}
\end{table*}

\begin{table*}[t]
    \centering
    \caption{For ORB, SuperPoint, and PX4FLOW, in combination with our template pipeline, the relative translation errors on the indoor sequences for randomly sampled sub-trajectories of the indicated lengths are given. The pipeline configurations shown in \cref{tab:accuracy_results_rmse} were used.}
    \aboverulesep=0ex
    \belowrulesep=0ex
    \begin{tabular}{l||c|c|c|c||c|c|c|c}
        \toprule
        \textbf{Sequence} & \multicolumn{4}{c||}{\textbf{Mean Rel. Translation Error over 15\,m $\downarrow$}} & \multicolumn{4}{c}{\textbf{Mean Rel. Translation Error over 25\,m $\downarrow$}} \\ 
        \midrule
        & ORB & SuperPoint & PX4FLOW & Original PX4FLOW & ORB & SuperPoint & PX4FLOW & Original PX4FLOW \\
        \midrule
        02 & 30.0\% & \textbf{18.9\%} & 30.3\% & 58.6\% &  \textbf{0.9\%} &  6.1\% &  \textbf{0.9\%} & 24.2\% \\
        03 & 15.0\% & 23.4\% & \textbf{14.7\%} & 32.8\% &  \textbf{6.3\%} & 11.6\% &  6.4\% & 31.5\% \\
        04 & 12.2\% & 19.3\% & \textbf{11.6\%} & 19.2\% &  7.9\% & 18.8\% &  \textbf{7.7\%} & 16.8\% \\
        05 & 15.8\% & 25.4\% & 16.2\% & \textbf{15.4\%} &  5.8\% & 23.6\% &  5.6\% &  \textbf{4.5\%} \\
        06 & 34.3\% & 29.7\% & \textbf{28.8\%} & 46.9\% & 17.9\% & 16.5\% & 14.0\% & \textbf{13.8\%} \\
        07 & 11.6\% & 36.0\% &  \textbf{8.8\%} &  9.2\% & 14.4\% & 31.5\% & \textbf{12.0\%} & 12.4\% \\
        08 &  \textbf{9.1\%} & 25.6\% & 12.0\% & 17.5\% &  \textbf{4.8\%} & 20.6\% &  6.1\% &  9.3\% \\
        \bottomrule
    \end{tabular}
    \label{tab:accuracy_results_epe}
\end{table*}

Since we generate the SuperPoint feature extraction model using \emph{NNTool}, we do not have a single core implementation to compare against. For the post-processing and matching, we indicate the hypothetical parallelization speed-up. In reality, however, we dedicate all the cluster resources to the feature extraction stage and run the post-processing and matching on the fabric controller. The neural network model does not fit fully into the L2 memory and needs to be partially stored in off-chip L3 memory. The memory accesses impact the processing time and are reflected in the reported performance figures in the step \emph{SP - Feature Extraction}.

It is worth mentioning that the time complexity of PX4FLOW is quadratically dependent on the maximum trackable movement. Hence, the $\pm 4.5$ pixel limitations shown in \Cref{tab:accuracy_results_rmse} could be increased to $\pm 8.5$ pixels by quadrupling the computation time of PX4FLOW, as further discussed in \Cref{sec:ablation_px4flow}. On the contrary, the time complexities of ORB and SuperPoint are independent of the trackable displacement and hence the movement velocity. In the case of the ORB tracker, the computation time mainly depends on the quality of the texture. To keep the number of ORB features and hence the computation time roughly constant, we apply hystereses on the FAST and Harris thresholds, which target between 150 and 200 descriptors. We implement a hard upper bound on the number of descriptors of 512, as well as hard upper and lower bounds on the thresholds to handle feature-rich and low-texture scenarios, respectively.

When considering the cycle counts of the three pipelines in combination with the cycles required for the template pipeline steps and running GAP9 at 370\,MHz, we could run PX4FLOW at max. 1216.3\,FPS, ORB at 118.5\,FPS, and SuperPoint at 11.9\,FPS, respectively. Since the framerate of PX4FLOW is significantly above the 300\,FPS that our camera is recording with, the power consumption of GAP9 can be optimized by keeping the compute cluster idle in between frames or by only using two of the eight compute cluster cores or by reducing the clock speed to 95 MHz, resulting in 312\,FPS but also increasing latency. Using only two of the eight cluster cores will strike a better power-to-latency trade-off at a power draw of 35\,mW to 45\,mW \cite{kiamarzi2024qr}. The maximum achievable frame rate of 118.5\,FPS of ORB renders GAP9 an ideal platform for algorithms with a similar computational complexity, thanks to its multicore cluster and efficient architecture.

\begin{table*}[t]
    \centering
    \caption{The table shows the details of each recorded outdoor benchmarking sequence. For ORB, SuperPoint, and PX4FLOW, in combination with our template pipeline, the relative translation errors for randomly sampled sub-trajectories of the indicated lengths are given. The pipeline configurations shown in \cref{tab:accuracy_results_rmse} were used.}
    \aboverulesep=0ex
    \belowrulesep=0ex
    \begin{tabular}{l|l|c|c||c|c|c||c|c|c}
        \toprule
        \textbf{Sequence} & \textbf{Texture} & \textbf{Duration [s]} & \textbf{Length [m]} & \multicolumn{3}{c||}{\textbf{Mean Rel. Translation Error over 25\,m $\downarrow$}} & \multicolumn{3}{c}{\textbf{Mean Rel. Translation Error over 50\,m $\downarrow$}} \\ 
        \midrule
        & & & & ORB & SuperPoint & PX4FLOW & ORB & SuperPoint & PX4FLOW \\
        \midrule
        %1 Hardcourt & 103.5 &  85.8 & 31.2\% & 46.4\% &  \textbf{4.3\%} & 33.2\% & 36.0\% &  \textbf{3.2\%} \\
        %2 
        Hardcourt 1 & sparse & 152.2 & 120.9 & 32.4\% & 22.1\% & \textbf{9.6\%} & 37.2\% & 22.6\% &  \textbf{6.8\%} \\
        %3 
        Hardcourt 2 & sparse & 182.4 & 172.4 & 27.4\% & 54.7\% & \textbf{12.6\%} & 22.8\% & 57.3\% & \textbf{11.5\%} \\
        %4 Harcourt & 124.0 & 121.3 & 23.0\% & 31.4\% & \textbf{20.7\%} & \textbf{21.4\%} & 26.9\% & 23.0\% \\
        %5 
        Mixed & mixed & 120.0 &  99.3 & 26.8\% & 43.7\% & \textbf{13.0\%} & 25.3\% & 29.3\% & \textbf{8.2\%} \\
        %6 
        Grass & cluttered & 138.3 & 126.9 & \textbf{7.5\%} & 56.7\% & 15.0\% &  \textbf{5.9\%} & 58.6\% & 12.6\% \\
        %7 
        Sand & cluttered & 141.1 & 118.9 & \textbf{10.6\%} & 38.0\% & 12.4\% &  \textbf{6.8\%} & 31.1\% & 10.9\% \\
        \bottomrule
    \end{tabular}
    \label{tab:accuracy_results_epe_outdoor}
\end{table*}

\subsection{Accuracy}
We indicate the accuracy of the three feature trackers using our template pipeline, as well as the accuracy of the original PX4FLOW pipeline on our indoor benchmark dataset in \Cref{tab:accuracy_results_rmse} in terms of the RMSE and in \Cref{tab:accuracy_results_epe} in terms of the relative translation error. The relative translation errors on the outdoor dataset are given in \Cref{tab:accuracy_results_epe_outdoor}.

In \Cref{tab:accuracy_results_rmse}, we observe that the original PX4FLOW configuration works very well in pure translation scenarios (sequences 05 and 07), indicating that the inclusion of the rigid body motion estimation introduces additional inaccuracies in those situations. In contrast, the original PX4FLOW configuration struggles with large turns ($>90$ degrees), as shown by the proposed pipeline, in combination with ORB and PX4FLOW, outperforming it in all other sequences in terms of RMSE. We can deduce that adding the rigid-body motion model makes the pipeline less accurate in pure translation movements but increases the robustness of accurate tracking under general movements significantly.

Due to the 8x subsampling pattern of SuperPoint, outliers are more systematic than in PX4FLOW or ORB, making the statistical outlier rejection of our rigid-body motion estimation less effective. Therefore, SuperPoint performs inferiorly in our template pipeline in terms of RMSE since those systematic outliers cause errors in the rotation estimation of the rigid-body motion model. However, SuperPoint performs well on sub-trajectories, especially when little direction changes are present, which is the case for both \emph{Square} movement sequences (02 and 06), as can be seen in \Cref{tab:accuracy_results_epe}.

In the sequences with movement in a square or random movement, both ORB and PX4FLOW perform similarly well. The interpolation of half-pixel movements in PX4FLOW makes it competitive with the larger ORB and SuperPoint models but at the cost of a significantly shorter tracking range ($\pm 4.5$ pixels versus virtually arbitrary big movements). The small tracking range of PX4FLOW is especially problematic in fast turns, where the motion is small in the rotation center but grows the further a pixel is away from the rotation center. In feature-rich environments, like on the rug, this is less problematic, but in environments with no features in the rotation center, like in Sequences 06 and 08, this can lead to estimation errors.

\begin{figure}[ht]
    \centering
    \includegraphics[width=\linewidth]{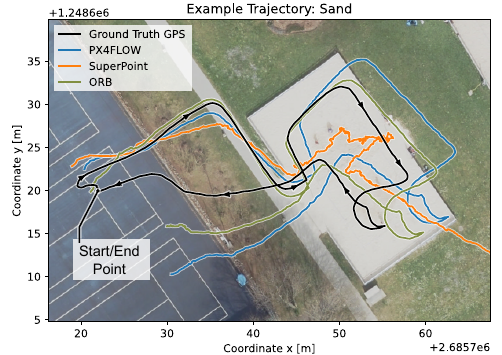}
    \caption{Real world example trajectory (Sand) including the ground truth, as well as the results of the feature trackers.}
    \label{fig:example_trajectory}
\end{figure}

The outdoor experiments presented in \Cref{tab:accuracy_results_epe_outdoor} and \Cref{fig:example_trajectory} demonstrate that the ORB and SuperPoint trackers struggle with the lack of textures on the hardcourt surface. The PX4FLOW feature tracker, which uses fixed pixel locations for the calculation of optical flow and does not require a minimum feature quality, performs best in the hardcourt and mixed sequences. The ORB tracker performs worse on the mixed sequence due to the hardcourt parts and performs best when on feature-rich textures like grass or sand.

\subsection{Ablation Study of PX4FLOW}
\label{sec:ablation_px4flow}
When comparing the original PX4FLOW pipeline with PX4FLOW in combination with our template pipeline in \Cref{tab:accuracy_results_rmse,tab:accuracy_results_epe}, we observe that the rigid-body motion assumption improves tracking accuracy. It is, however, worth noting that in the patch-based approach of PX4FLOW, the maximum tracking speed is heavily dependent on the frame rate. For a more complete comparison, we therefore conducted an ablation study of multiple PX4FLOW configurations.

As can be seen in \Cref{tab:px4flow_ablation_results}, only increasing the frame rate to 300\,FPS while leaving the template pipeline unchanged leads to inferior performance in all sequences. The template pipeline would need to be tuned towards the higher frame rate, which includes adapting the outlier thresholds of the rigid body motion estimation, as well as tuning the gains of the Kalman filter.

\begin{figure}[h]
    \centering
    \includegraphics[width=\linewidth]{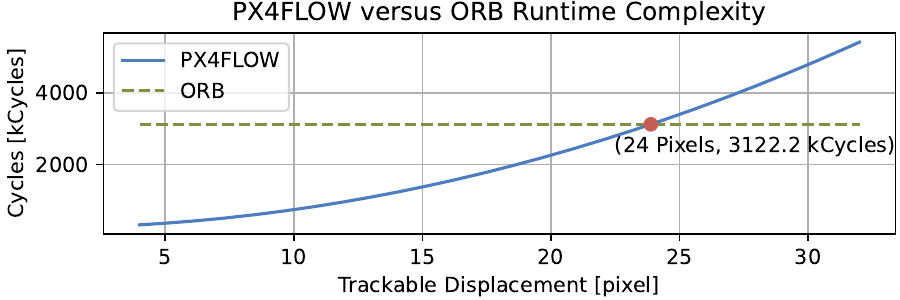}
    \caption{The runtime complexities of the template pipeline in combination with PX4FLOW and ORB, depending on the maximum displacement that shall be tracked between two frames.}
    \label{fig:runtime_orb_px4flow}
\end{figure}

A more reliable method to increase the trackable speed (i.e., the trackable distances per frame) of PX4FLOW is to double the pixel search range to $\pm 8.5$ while keeping the framerate at 100\,FPS. Due to the runtime complexity being quadratic with respect to the search range, doubling the trackable distance quadruples the required cycle count. Given the competitive accuracy of PX4FLOW and the short runtime as determined in \Cref{sec:results_profiling_feature_trackers}, this can be an interesting option, as the PX4FLOW variant with $\pm 8.5$ pixel search range outperforms the original algorithm in 4 sequences. Given the scaling properties of PX4FLOW, it is, however, not possible to reach the arbitrary tracking ranges of SuperPoint and ORB. When increasing the tracking range of PX4FLOW by a factor of 6 to 24.5 pixels, we reach the same runtime as for ORB, which is set to track displacements of 32 pixels, a value that can be increased without any impact on the runtime as shown in \Cref{fig:runtime_orb_px4flow}.

\begin{table}[t]
    \centering
    \caption{Ablation study of varying PX4FLOW configurations in combination with our template pipeline. The table shows the RMSE depending on framerate and maximum trackable movement distance in pixels.}
    \aboverulesep=0ex
    \belowrulesep=0ex
    \begin{tabular}{l|c|c|c}
        \toprule
        Framerate & 100 FPS & 100 FPS & 300 FPS \\
        Max Movement & $\pm 4.5$ pixel & $\pm 8.5$ pixel & $\pm 4.5$ pixel \\
        \midrule
        \midrule
        \textbf{Sequence} & \multicolumn{3}{c}{\textbf{RMSE [m]} $\downarrow$ and (Standard Deviation)} \\
        \midrule
        02 & \textbf{0.275} (0.145) & 0.325 (0.180) & 1.132 (0.658) \\
        03 & \textbf{0.394} (0.216) & 0.450 (0.251) & 2.545 (1.396) \\
        04 & 0.464 (0.262) & \textbf{0.367} (0.210) & 1.347 (0.746) \\
        05 & 0.320 (0.192) & \textbf{0.290} (0.177) & 3.582 (2.234) \\
        06 & 1.631 (0.984) & \textbf{1.228} (0.765) & 1.910 (0.978) \\
        07 & \textbf{0.554} (0.296) & 0.728 (0.346) & 3.673 (2.266) \\
        08 & 1.130 (0.574) & \textbf{0.553} (0.257) & 1.427 (0.853) \\
        \bottomrule
    \end{tabular}
    \label{tab:px4flow_ablation_results}
\end{table}

\section{Discussion}
\label{sec:discussion}
The results show simplifying VIO to a downfacing configuration is a valid approach to reduce the computational demand while achieving accurate VIO predictions. Due to the restriction to a downfacing camera configuration and the assumption of planar motion, the VIO pipeline variants presented in this work are mainly suitable for structured environments like indoor spaces or cities. While those two requirements restrict the use of the presented VIO solutions, we consider it a first step towards realizing an accurate VIO pipeline on efficient hardware. The GAP9 SoC used throughout this work consumes less than 68\,mW \cite{kiamarzi2024qr, cereda2024device} when running at 370\,MHz. This is two orders of magnitude lower than comparable implementations on Raspberry Pi class devices \cite{bahnam2021stereo} and in the same order of magnitude as the application-specific integrated circuit implementation Navion, which uses 24\,mW to process frames at 171\,FPS \cite{suleiman2019navion}.

The three assessed feature trackers have different strengths and weaknesses. The modified PX4FLOW variant shown in this work is computationally still very efficient (148.2\,kCycles for the motion prediction before fusion with the IMU versus 88.8\,kCycles in the simpler parallelized version in \cite{kuhne2022parallelizing}) while surpassing the accuracy of the original PX4FLOW implementation on five of the seven sequences of the indoor dataset. However, it shows limited scalability to faster movement velocities ($\geq$ 24 pixels per frame) or higher image resolutions due to the quadratic scaling of computational load in terms of maximum trackable displacement per frame, as shown in \Cref{fig:runtime_orb_px4flow}. The ORB-based variant strikes a favorable trade-off between trackable displacement, which can be arbitrary due to feature descriptors, and computational load, which is ten times higher compared to PX4FLOW due to the added feature detection, description, and matching logic. Lastly, SuperPoint is competitive in terms of accuracy, but the achievable frame rate of 11.9\,FPS needs improvement. It is important to note that SuperPoint is mainly memory-bound, as the full model does not fit into L2 and L1 memory and needs to be partially stored on off-chip L3 memory. Furthermore, SuperPoint has been heavily quantized from 32-bit floating point to 8-bit fixed point values, resulting in a mean localization error of 1.71 pixels on the detector and reducing the cosine similarity of the descriptors to 0.91 with respect to the full precision model. The use of a SoC with large enough on-chip memory could render SuperPoint a competitive option.

Depending on the available computational resources and requirements regarding trackable movement velocities, the middle ground between the PX4FLOW and ORB feature trackers could be examined further. Instead of using fixed points of interest, a PX4FLOW derivative could use a feature detector like FAST \cite{rosten2006machine} akin to ORB. Alternatively, an ORB derivative could use a simpler feature detector and/or descriptor. Furthermore, an ORB derivative could employ a subpixel refinement similar to that of PX4FLOW.

% BiomedBench https://ieeexplore.ieee.org/abstract/document/10720867 (14.6mW)
% QR-PULP https://dl.acm.org/doi/pdf/10.1145/3649153.3649210
% (166887 0.013mJ, 68376 0.012mJ, 1168717 0.216mJ) -> 28 mW, 64mW, 68 mW
% Palossi paper: https://arxiv.org/pdf/2403.04071 (66mW)
% Navion: https://ieeexplore.ieee.org/abstract/document/8600375
% 2 mW at 20 fps and 24mW at 171 fps

\section{Conclusion}
\label{sec:conclusion}
In this work, we evaluate the viability of downfacing VIO when using various feature trackers (PX4FLOW, ORB, and SuperPoint) on resource-constrained SoCs, suitable for micro- and nano-drones. Furthermore, we present a template VIO pipeline suitable for modern RISC-V-based architectures, which yields an accuracy improvement in RMSE by a factor of 3.65x over previous microcontroller implementations. We show that for smaller movements, PX4FLOW in our modified version is still a valid choice to this date, whereas, for larger movements above 24 pixels/frame, our ORB pipeline yields accurate results. For future work, we deem ORB, in combination with the subpixel refinement of PX4FLOW, an interesting combination for VIO computation on drones with strict resource constraints.

\bibliographystyle{IEEEtran}
\bibliography{main}

\begin{IEEEbiography}[{\includegraphics[width=1in,height=1.25in,clip,keepaspectratio]{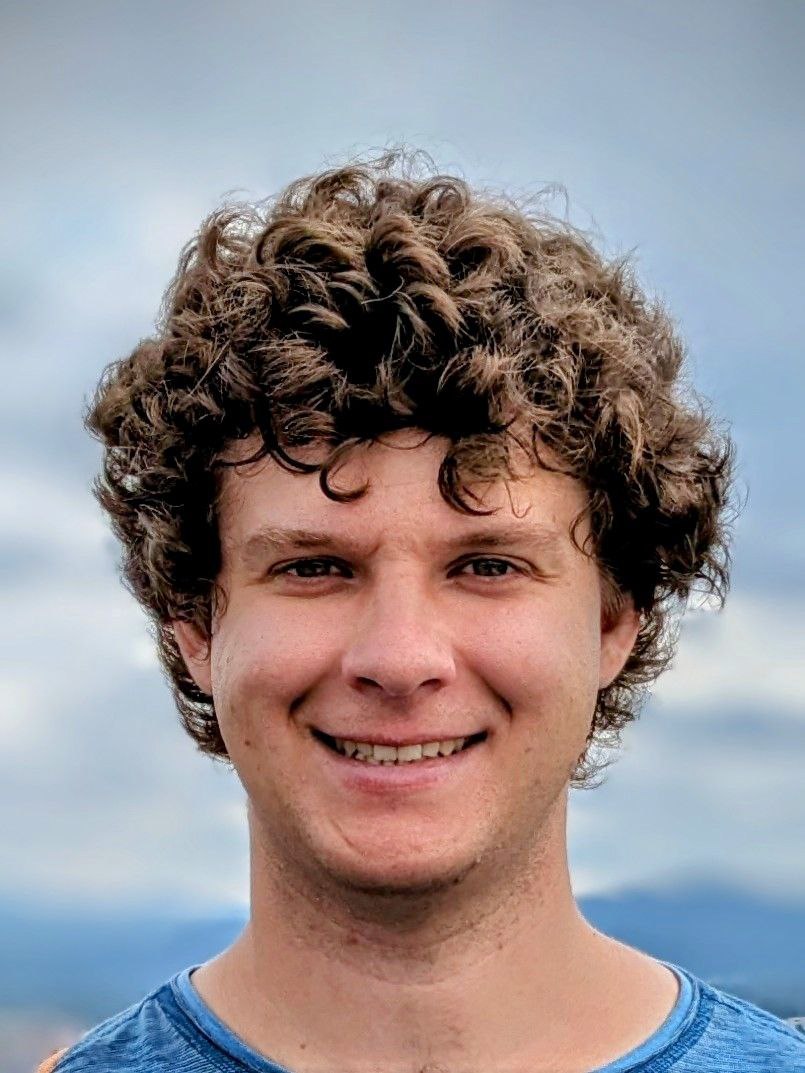}}]{Jonas Kühne}
(Graduate Student Member, IEEE) received the B.Sc. and M.Sc. degrees in electrical engineering and information technology from ETH Zürich, Zürich, Switzerland, in 2016 and 2018, respectively. Between 2019 and 2021, he worked for Agtatec AG, which is part of Assa Abloy. \\
He is currently pursuing his Ph.D. degree with both the Integrated Systems Laboratory and the D-ITET Center for Project-Based Learning at ETH Zürich, Zürich, Switzerland. \\ 
His research interests include algorithm and hardware design for visual inertial odometry and SLAM on low-power embedded systems. 
\end{IEEEbiography}

\begin{IEEEbiography}[{\includegraphics[width=1in,height=1.25in,clip,keepaspectratio]{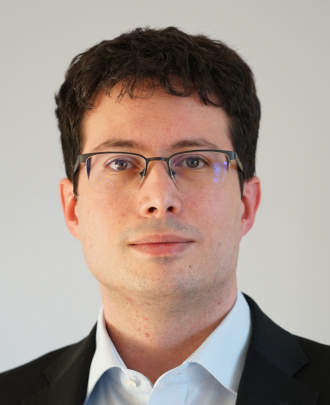}}]{Christian Vogt} (Member, IEEE) received the M.Sc. degree and the Ph.D. in electrical engineering and information technology from ETH Zürich, Zürich, Switzerland, in 2013 and 2017, respectively. He is currently a post-doctoral researcher and lecturer at ETH Zürich, Zürich, Switzerland. His research work focuses on signal processing for low power applications, including field programmable gate arrays (FPGAs), IoT, wearables and autonomous unmanned vehicles.
\end{IEEEbiography}

\begin{IEEEbiography}[{\includegraphics[width=1in,height=1.25in,clip,keepaspectratio]{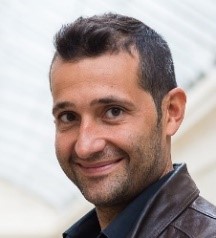}}]{Michele Magno}
(Senior Member, IEEE) received his master's and Ph.D. degrees in electronic engineering from the University of Bologna, Bologna, Italy, in 2004 and 2010, respectively. \\
Currently, he is a \emph{Privatdozent} at ETH Zurich, Zurich, Switzerland, where he is the Head of the Project-Based Learning Center. He has collaborated with several universities and research centers, such as Mid University Sweden, where he is a Guest Full Professor. He has published more than 150 articles in international journals and conferences, in which he got multiple best paper and best poster awards. The key topics of his research are wireless sensor networks, wearable devices, machine learning at the edge, energy harvesting, power management techniques, and extended lifetime of battery-operated devices.
\end{IEEEbiography}

\begin{IEEEbiography}[{\includegraphics[width=1in,height=1.25in,clip,keepaspectratio]{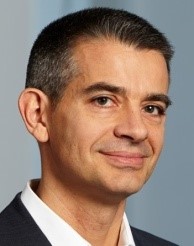}}]{Luca Benini}
(Fellow, IEEE) received the Ph.D. degree in electrical engineering from Stanford University, Stanford, CA, USA, in 1997.\\
He holds the Chair of Digital Circuits and Systems at ETH Zurich, Zurich, Switzerland, and is a Full Professor at the University of Bologna, Bologna, Italy. His current research interests include energy-efficient computing systems' design from embedded to high performance.\\
Dr. Benini is a fellow of the ACM and a member of the Academia Europaea. He was a recipient of the 2016 IEEE CAS Mac Van Valkenburg Award and the 2023 McCluskey  Award.
\end{IEEEbiography}

\vfill

\end{document}